\documentclass[11pt, oneside]{article}    

\usepackage{geometry}
\usepackage{graphicx}
\usepackage{color}
\usepackage{xcolor}
\usepackage{amssymb}
\usepackage{hyperref}
\usepackage{amsmath,amsfonts,bm,bbm}

\def\eqref#1{equation~\ref{#1}}
\def\1{\bm{1}}

\def\rvw{{\mathbf{w}}}
\def\rvx{{\mathbf{x}}}

\def\rmW{{\mathbf{W}}}

\DeclareMathAlphabet{\mathsfit}{\encodingdefault}{\sfdefault}{m}{sl}
\SetMathAlphabet{\mathsfit}{bold}{\encodingdefault}{\sfdefault}{bx}{n}

\def\gH{{\mathcal{H}}}
\def\gI{{\mathcal{I}}}
\def\gJ{{\mathcal{J}}}

\def\gO{{\mathcal{O}}}

\def\sR{{\mathbb{R}}}

\usepackage{mathrsfs}
\usepackage{caption}
\usepackage{subcaption}

\title{NeurIPS  2020 Competition:\\ Predicting Generalization in Deep Learning ({\color{red}Version 1.1})}
\author{Yiding Jiang \thanks{Lead organizer: Yiding Jiang; Scott Yak and Pierre Foret help implement large portion of the infrastructure and the remaining organizers' order is randomized.} \thanks{Affiliation: Google Research}\and Pierre Foret\footnotemark[2] \and Scott Yak\footnotemark[2] \and Daniel M. Roy\thanks{Affiliation: University of Toronto}  \thanks{Equal contribution: random order} \and Hossein Mobahi\footnotemark[2] \footnotemark[4]  \and Gintare Karolina Dziugaite\thanks{Affiliation: Element AI} \footnotemark[4] \and Samy Bengio\footnotemark[2] \footnotemark[4] \and Suriya Gunasekar\thanks{Affiliation: Microsoft Research} \footnotemark[4] \and Isabelle Guyon \thanks{Affiliation: University Paris-Saclay and ChaLearn} \footnotemark[4] \and Behnam Neyshabur\footnotemark[2] \footnotemark[4] \and\\
{\color{magenta}\texttt{\texttt{pgdl.neurips@gmail.com}}}
}
\date{\today}

\begin{document}

\newcommand{\btheta}[1]{$\boldsymbol{\theta}$}
\newcommand{\prompt}[1]{}

\maketitle

\begin{abstract}
% Briefly describe your competition, either a ``regular competition'' running over a few months before the challenge or a live/demonstration competition (competition of demonstrations or live contest) running at the conference site.
% Summarize the background, available data, methods, available baseline, and potential impact.
Understanding generalization in deep learning is arguably one of the most important questions in deep learning. Deep learning has been successfully adopted to a large number of problems ranging from pattern recognition to complex decision making, but many recent researchers have raised many concerns about deep learning, among which the most important is \textbf{generalization}. Despite numerous attempts, conventional statistical learning approaches have yet been able to provide a satisfactory explanation on why deep learning works. A recent line of works aims to address the problem by trying to predict the generalization performance through complexity measures. In this competition, we invite the community to propose complexity measures that can accurately predict generalization of models. A robust and general complexity measure would potentially lead to a better understanding of deep learning's underlying mechanism and behavior of deep models on unseen data, or shed light on better generalization bounds. All these outcomes will be important for making deep learning more robust and reliable.
\end{abstract}

\subsection*{Keywords}
Generalization, Deep Learning, Complexity Measures
\subsection*{Competition type} Regular.

\section{Competition description}

\subsection{Background and impact}

% Generalization of deep neural network is one of the most important unanswered questions in deep learning. Despite the its empirical success, both the theoreticians and machine learning practitioners still do not have a convincing understanding of why these over-parameterized models work.

% Generalization.

% Architecture search.

% Climate change.

\prompt{Provide some background on the problem approached by the competition and fields of research involved. Describe the scope and indicate the anticipated impact of the competition prepared (economical, humanitarian, societal, etc.). \textbf{Please note that tasks of humanitarian and/or positive societal impact will be particularly considered this year.}}

Deep learning has been successful in a wide variety of tasks, but a clear understanding of underlying root causes that control generalization of neural networks is still elusive. A recent empirical study \cite{jiang2019fantastic} looked into many popular complexity measures. By a carefully controlled analysis on hyperparameter choices being a confounder for both generalization and the complexity measure, they came to surprising findings about which complexity measures worked well and which did not. However, rigorously evaluating these complexity measures required training many neural networks, computing the complexity measures on them, and analyzing statistics that condition over all variations in hyperparameters. Such cumbersome process makes it painstaking, error-prone, and computationally expensive. As the results, this procedure is not accessible to members of the wider machine learning community who do not have access to larger compute power.

By hosting this competition, we intend to provide a platform where participants only need to write the code that computes the complexity measure for a trained neural network, and let the competition evaluation framework handle the rest. This way, participants can focus their efforts on coming up with the best complexity measure instead of replicating the experimental set up. In addition, the ML community benefits from results that are directly comparable with each other, which alleviates the need for having every researcher to reproduce all the benchmark results themselves.

\prompt{Justify the relevance of the problem to the targeted by the NeurIPS community and indicate whether it is of interest to a large audience or limited to a small number of domain experts (estimate the number of participants). A good consequence for a competition is to learn something new by answering a scientific question or make a significant technical advance.}

This problem is likely to be of interest to machine learning researchers who study generalization of deep learning or experts in learning theory, and the neural architecture search community. We hope that this competition would enable researchers to quickly test theories of generalization with a shorter feedback loop, thereby leading to stronger foundations for designing high-performance, efficient, and reliable deep learning algorithms/architectures. In addition, the fact that all proposed approaches are assessed within the same evaluation system can ensure a fair and transparent evaluation procedure.

\prompt{
Describe typical real life scenarios and/or delivery vehicles for the competition. This is particularly important for live competitions, but may also be relevant to regular competitions. For instance: what is the application setting, will you use a virtual or a game environment, what situation(s)/context(s) will participants/players/agents be facing?}

This competition is unlike a typical supervised-learning competition -- participants are \emph{not} given a large and representative training set for training a model to produce predictions on an unlabelled test set. Instead, participants submit code, which would run on our evaluation server, and the results would be published on a public leaderboard updated on a daily basis. They are given a small set of models for debugging their code, but this set is not expected to be sufficient for training their model. Their code is expected to take a training dataset of image label pairs as well as a model fully trained on it as input, and generate a real number as output. \textit{The value of the output should ideally be larger for models that have larger generalization gaps.}

\prompt{
Put special emphasis on relating the, necessarily simplified, task of the competition to a real problem faced in industry or academia. If the task cannot be cast in those terms, provide a detailed hypothetical scenario and focus on relevance to NeurIPS.}

Generalization is one the most fundamental question of machine learning. A principled understanding of generalization can provide theoretical guarantees for machine learning algorithms, which makes deep learning more accountable and transparent, and is also desirable in safety critical applications. For example, generalization to different environments and data distribution shifts is a critical aspect for deploying autonomous vehicles in real life. The result of this competition could also have implications for more efficient architecture search which could reduce the carbon footprint of designing machine learning models and have environmental impact in the long run.

\subsection{Novelty}

\prompt{Have you heard about similar competitions in the past? If yes, describe the key differences.
Indicate whether this is a completely new competition, a competition part of a series, eventually re-using old data.}

This competition is quite unique, and there are no previous competitions similar to it. The competition is focused on achieving better theory and understanding of the observed phenomena. We hope the competition will allow the fair comparison of numerous theories the community has proposed about generalization of deep learning; however, we also welcome the broader data science community to find practical solutions that are robust under the covariate shift in data and architecture, but do not necessarily give rise statistical bounds or direct theoretical analysis.

\subsection{Data}

\prompt{If the competition uses an evaluation based on the analysis of data,
please provide detailed information about the available data and
their annotations, as well as permissions or licenses to use such data. If new data are collected or generated, provide details on the procedure, including permissions to collect such data obtained by an ethics committee, if human subjects are involved. In this case, it must be clear in the document that the data will be ready prior to the official launch of the
competition. Please justify that: (1) you have access to large
enough datasets to make the competition interesting and draw
conclusive
 results; (2) the data will be made freely available;(3) the ground truth has been kept confidential.}

As previously outlined, our competition differs from the traditional supervised learning setting, because our data-points are trained neural networks. As such, the dataset provided to competitors will be a \textit{collection of trained neural networks}. We chose the Keras API (integrated with TensorFlow 2.x) for its ease of use and the familiarity that a large part of the community has developed with it. Furthermore, all our models will be sequential (i.e., no skip-connections), making it more intuitive for the competitors to explore the architecture of each network. To remove a potential confounding factor, all models are trained until they reach the interpolation regime (close to 100\% accuracy on the training set \footnote{Alternative, we can also consider margin based stopping criterion (i.e. difference between the highest logit and second highest logit).} or to a fixed cross entropy value). Each model is represented by a JSON file describing its architecture and a set of trained weights (HDF5 format).  We provide the helper functions needed to load a model from those files.

\begin{figure}[h]
  \begin{subfigure}{.5\textwidth}
  \centering
  \includegraphics[width=.9\linewidth]{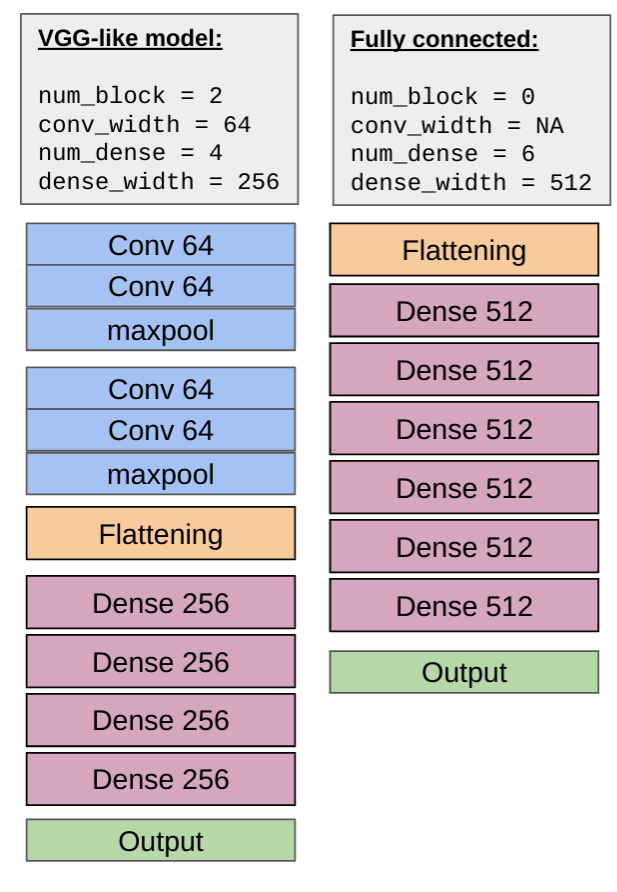}
  \caption{Architecture followed by VGG-like models.}
  \label{fig:arch_def}
\end{subfigure}%
\begin{subfigure}{.5\textwidth}
  \centering
  \includegraphics[width=.9\linewidth]{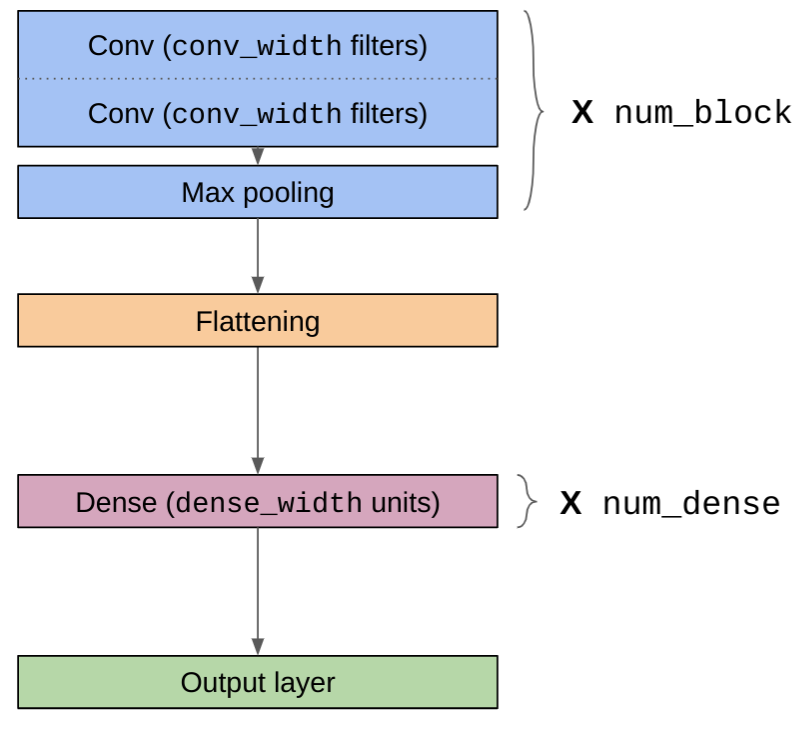}
  \caption{Architectures for two sets of VGG-like hyper parameters.}
  \label{fig:sub1}
\end{subfigure}%
\caption{Example architectures of the networks in our dataset.}
\end{figure}

The first architecture type we consider is derived from the parameterized architecture described in Figure \ref{fig:arch_def}. Due to its similarity to the VGG model \cite{Simonyan15}, we will refer to it as \textit{VGG-like models}. The second connectivity pattern we consider resembles \cite{lin2013network} (referred later as \textit{Network in network} which is generally more parameter efficient than VGG models due to the usage of global average pooling yet yields competitive performance. 

\begin{itemize}
    \item These types of models are well represented in the literature.  Inspired by the seminal work of \cite{Simonyan15}, they made their way into an important body of theoretical research in deep learning, see for instance \cite{2018arXiv180303635F, 2018arXiv181005270L, 2015arXiv150602626H}; however, the exact nature of their generalization is still not well understood.
    \item Empirically, these models reach the interpolating regime quite easily for a large number of optimizers / regularization, while exhibiting a large range of out of sample accuracy (sometime up to 20\% of test accuracy difference, for two model with the same architecture and regularization, both reaching 100\% training accuracy, depending on the optimizer / learning rate / batch size). In other words, we select these architectures as they often exhibit large generalization gaps and are thus well-suited for this competition.
    \item These architectures keep the number of hyper-parameters reasonable while exploring a large range of neural network types (fully connected or convolutional, deep or shallow, with convolutional / dense bottlenecks or not), although future iterations of the competition may include arbitrary type of computational graph.
\end{itemize}

It is worth noting that the optimization method chosen will not be visible to the competitor (i.e. dropout layers will be removed and optimizer is not an attribute of the model), although certain aspects of the hyperparameters can be inferred from the models (e.g. depth or width).

\subsubsection{Public and private tasks}

The competition is composed of the the following phases:

\begin{itemize}

\item Phase 0: Public data was given to the competitors: they were able to download Task~1 and Task 2 to measure the performance of their metrics locally, before uploading submissions to our servers.
\item Phase 1: First online leaderboard, accessible at the beginning of the competion (also called \textit{public leaderboard}). This leaderboard is composed of Task 4 and Task 5 and was used to compute the scores displayed on the leaderboard for the first phase of the competition. There was no Task 3 released in this competition, but we keep the original numbering of the tasks to avoid any confusion.
\item Phase 2: Private leaderboard, only accessible in the last phase of the competition, where competitors can upload their very best metrics. Winners are determined only on their score on this leaderboard (to prevent overfitting of the public leaderboard, as usual). This phase is composed of Task 6, Task 7, Task 8, and Task 9.
\end{itemize}

\textbf{Task 1}

\begin{itemize}
\item Model: VGG-like models, with 2 or 6 convolutional layers [conv-relu-conv-relu-maxpool] x 1 or x 3.  One or two dense layers of 128 units on top of the model. When dropout is used, it is added after each dense layer.
\item Dataset: CIFAR-10 \cite{cifar10}(10 classes, 3 channels).
\item Training: Trained for at most 1200 epochs, learning rate is multiplied by 0.2 after 300, 600 and 900 epochs. Cross entropy and SGD with momentum 0.9. Initial learning rate of 0.001
\item Hparams: Number of filters of the last convolutional layer in [256, 512]. Dropout probability in [0, 0.5]. Number of convolutional blocks in [1, 3]. Number of dense layers (excluding the output layer) in [1, 2]. Weight decay in [0.0, 0.001]. Batch size in [8, 32, 512].
 \end{itemize}

\textbf{Task 2}

\begin{itemize}
\item Model: Network in Network. When dropout is used, it is added at the end of each block.
\item Dataset: SVHN \cite{svhn} (10 classes, 3 channels)
\item Training: Trained for at most 1200 epochs, learning rate is multiplied by 0.2 after 300, 600 and 900 epochs. Cross entropy and SGD with momentum 0.9. Initial learning rate of 0.01.
\item Hparams: Number of convolutional layers in [6, 9, 12], dropout probability in [0.0, 0.25, 0.5], weight decay in [0.0, 0.001], batch size in [32, 512, 1024].
\end{itemize}

\textbf{Task 4}

\begin{itemize}
\item Model: Fully convolutional with no downsampling. Global average pooling at the end of the model. Batch normalization (pre-relu) on top of each convolutional layer.
\item Dataset: CINIC-10 \cite{cinic10} (random subset of 40\% of the original dataset).
\item Training: Trained for at most 3000 epochs, learning rate is multiplied by 0.2 after 1800, 2200 and 2400 epochs. Initial learning rate of 0.001.
\item Hparams: Number of parameters in [1M, 2.5M], Depth [4 or 6 conv layers], Reversed [True of False]. If False, deeper layers have more filters. If True, this is reversed and the layers closer to the input have more filters. Weight decay in [0, 0.0005]. Learning rate in [0.01, 0.001]. Batch size in [32, 256].
\end{itemize}

\textbf{Task 5}
\begin{itemize}
\item[] Identical to Task 4 but without batch normalization.
\end{itemize}

\textbf{Task 6}

\begin{itemize}
\item Model: Network in networks.
\item Dataset: Oxford Flowers \cite{oxfordflowers}(102 classes, 3 channels), downsized to 32 pixels.
\item Training: Trained for 10000 epochs maximum,  learning rate is multiplied by 0.2 after 4000, 6000, and 8000 epochs. Initial learning rate: 0.01. During training, we randomly apply data augmentation (the cifar10 policy from AutoAugment) to half the examples.
\item Hparams: Weight decay in [0.0, 0.001], batch size in [512, 1024], number of filters in convolutional layer in [256, 512], number of convolutional layers in [6, 9, 12], dropout probability in [0.0, 0.25], learning rate in [0.1, 0.01].
\end{itemize}

\textbf{Task 7}

\begin{itemize}
\item Model: Network in networks, with dense layer added on top on the global average pooling layer. Example: For a NiN with 256-256 dense, the global average pooling layer will have an output size of 256, and another dense layer of 256 units is added on top of it before the output layer.
\item Dataset: Oxford pets \cite{oxfordpets} (37 classes, downsized to 32 pixels, 4 channels). During training, we randomly apply data augmentation (the CIFAR-10 policy from AutoAugment) to half the examples.
\item Training: Trained for at most 5000 epochs, learning rate is multiplied by 0.2 after 2000, 3000 and 4000 epochs. Initial learning rate of 0.1.
\item Hparams: Depth in [6, 9], dropout probability in [0.0, 0.25], weight decay in [0.0, 0.001], batch size in [512, 1024], dense architecture in [128-128-128, 256-256, 512]
\end{itemize}

\textbf{Task 8}

\begin{itemize}
\item Model: VGG-like models (same as in task1) with one hidden layer of 128 units.
\item Dataset: Fashion MNIST \cite{fashionmnist} (28x28 pixels, one channel).
\item Training: Trained for at most 1800 epochs, learning rate is multiplied by 0.2 after 300, 600 and 900 epochs. Initial learning rate of 0.001. Weight decay of 0.0001 applied to all models.
\item Hparams: Number of filters of the last convolutional layer in [256, 512]. Dropout probability in [0, 0.5]. Number of convolutional blocks in [1, 3]. Learning rate in [0.001, 0.01]. Batch size in [32, 512].
\end{itemize}

\textbf{Task 9}

\begin{itemize}
\item Model: Network in Network.
\item Dataset: CIFAR-10, with the standard data augmentation (random horizontal flips and random crops after padding by 4 pixels.
\item Training: Trained for at most 1200 epochs, learning rate is multiplied by 0.2 after 300, 600 and 900 epochs. Cross entropy and SGD with momentum 0.9. Initial learning rate of 0.01.
\item Hparams: Number of filters in the convolutional layers in [256, 512], Number of convolutional layers in [9, 12], dropout probability in [0.0, 0.25], weight decay in [0.0, 0.001], batch size in [32, 512].
\end{itemize}

\subsection{Tasks and application scenarios}

\prompt{Describe the tasks of the competition and explain to which specific real-world scenario(s) they correspond to. If the competition does not lend itself
to real-world scenarios, provide a justification. Justify that the problem posed are scientifically or technically challenging but not impossible to
solve. If data are used, think of illustrating the same scientific problem using several datasets from various application domains.}

In this competition, each dataset $\mathscr{D}$ has a set of training data $\mathscr{D}_{train}=\{(\rvx_i, y_i)\}_{i=i}^{N_t}$ and a set of validation data $\mathscr{D}_{val}=\{(\rvx_i, y_i)\}_{i=i}^{N_v}$. Further, it has a set of parameterized models uniquely identified by the models' hyperparameters, $\Theta = \{\boldsymbol{\theta}_0, \dots, \boldsymbol{\theta}_{n}\}$. Each hyperparameter  produces one set of model weights\footnote{Due to the stochasticity in SGD or in the initialization, the same set of hyperparameters can yield models that generalize differently. However, a good generalization measure should also be able to rank between these models too.}. The set of weights produced by $\Theta$ is $\rmW = \{\rvw_0, \dots, \rvw_n\}$ where $\rvw_i$ represents parameters of the model with $\boldsymbol{\theta}_i$ trained on $\mathscr{D}_{train}$.
We further denote the resulting model to be $f_{\rvw_i}$ and use $f_{\rvw_i}(\rvx_j)$ as the prediction of $\rvx_j$. The generalization gap of the model is formally defined as
\begin{equation}
    g(f_{\rvw}; \mathscr{D}) = \frac{1}{|\mathscr{D}_{val}|} \sum_{(\rvx, y) \in \mathscr{D}_{val}} \mathbbm{1}(f_{\rvw}(\rvx) \neq y) - \frac{1}{|\mathscr{D}_{train}|} \sum_{(\rvx, y) \in \mathscr{D}_{train}} \mathbbm{1}(f_{\rvw}(\rvx) \neq y),
\end{equation} where $\mathbbm{1}$ is the indicator function.
A complexity measure $\mu: (f_\rvw, \mathscr{D}_{val}) \rightarrow \sR$ maps the model and dataset to a real number. The task is to find a complexity measure $\mu$ such that:
\begin{equation}
\mathrm{sgn}(\mu(\rvw; \mathscr{D}_{train}) - \mu(\rvw'; \mathscr{D}_{train})) = \mathrm{sgn}(g(\rvw; \mathscr{D}) - g(\rvw'; \mathscr{D})) \quad \forall \,\, (\rvw, \rvw') \in \rmW  \times \rmW 
\end{equation}
Informally, the function $\mu$ should \textbf{order} the models in the same way that the generalization gap does. Further, for notation simplicity the dependency on $\mathscr{D}_{val}$ will be omitted unless discussing about multiple datasets.

In statistical learning theory, $\mu$ is often the upperbound of the generalization error a function can make; however, such complexity measure is often much larger than the admissible error made by deep neural networks with large number of parameters, rendering them vacuous. In these cases, these complexity measures can still be informative so long as they provide comparison between different models. Beyond the theoretical interests and usage in model selections, these complexity measures can also be instrumental in Neural Architecture Search (NAS) by alleviating the need of having a validation dataset. This can be critical in regimes where data are scarce and using valuable data for model selection is sub-optimal. Finally, if the complexity measure is fully differentiable, it may act as a regularizer to improve generalization.

The largest challenge of designing such complexity measure measure is making it robust to changes in the model architectures and the dataset used for training the model. Many complexity measures only work on a single dataset or only correlates with one particular type of hyper-parameter change (e.g. depth). While such a generic complexity measure may seem to be difficult to obtain, recent work \cite{jiang2019fantastic} has proposed rigorous protocols for identifying promising complexity measure, and shows that it is possible to find complexity measures that fulfill the above criteria.

\subsection{Metrics}

\prompt{For quantitative evaluations, select a scoring metric and justify
that it effectively assesses the efficacy of solving the problem
at hand. It should be possible to evaluate the results
objectively. If no metrics are used, explain how the evaluation
will be carried out. Explain how error bars will be computed and/or how the significance in performance difference between participants will be evaluated.}

In this competition, each dataset has a set of hyperparameters that are adjusted to create different models. Formally, we denote each hyperparameter by $\theta_i$ taking values from the set $\Theta_i$, for $i=1,\dots,n_H$ and $n_H$ denoting the total number of hyperparameter types. Each value of hyperparameters $\boldsymbol{\theta} \triangleq (\theta_1,\theta_2,\dots,\theta_{n_H}) \in \Theta$ is drawn from $\Theta \triangleq \Theta_1 \times \Theta_2 \times \dots \times \Theta_{n_H}$. For any pair of $(\theta, \theta') \in \Theta \times \Theta$, we define:
\begin{equation}
    V_\phi(\theta, \theta') \triangleq \mathrm{sgn}(\phi(\theta) - \phi(\theta'))
\end{equation}
Then the Kendall's ranking correlation between a measure and generalization gap is defined as:
\begin{equation}
    \tau(\mu, \Theta) \triangleq \frac{1}{|\Theta|(|\Theta|-1)} \sum_{\theta_1} \sum_{\theta_2 \ne \theta_1} V_\mu(\theta_1, \theta_2)V_g(\theta_1, \theta_2)
\end{equation}

\subsubsection{Metric 1: Controlled Ranking Correlation}
To ensure that the ranking correlation is good at capturing changes in every single hyperparameter; therefore, we design the first metric to reflect the measure ability to predict changes in generalization gap as the result of changes in a single hyperparameter:
% \begin{equation}
%     \psi(\mu, i) \triangleq \frac{1}{|\Theta|(|\Theta_i|-1)} \sum_{\theta \in |\Theta|} \sum_{\theta' \in |\Theta_i|} V_\mu(\theta, \theta_{\theta_i=\theta'})V_\mu(\theta, \theta_{\theta_i=\theta'})
% \end{equation}
{\small
\begin{equation}
m_i \triangleq |\Theta_1 \times \dots \times \Theta_{i-1} \times \Theta_{i+1} \times \dots \times \Theta_{n_H}|
\end{equation}
\begin{equation}
\psi(\mu, i) \triangleq \frac{1}{m_i} \sum_{\theta_1 \in \Theta_1} \dots \sum_{\theta_{i-1} \in \Theta_{i-1}} \sum_{\theta_{i+1} \in \Theta_{i+1}} \dots \sum_{\theta_{n_H} \in \Theta_{n_H}} \tau\,(\mu, \cup_{\theta_i \in \Theta_i} \{ (\theta_1,\dots,\theta_i, \dots, \theta_{n_H}) \})
\end{equation}
The inner $\tau$ reflects the ranking correlation between the generalization and the complexity measure for a small group of models where the only difference among them is the variation along a single hyperparameter $\theta_i$. We then average the value across all combinations of the other hyperparameter axis. Intuitively, if a measure is good at predicting the effect of hyperparameter $\theta_i$ over the model distribution, then its corresponding $\psi_i$ should be high. Finally, we compute the average $\psi_i$ across all hyperparamter axes, and name it $\Psi$
\begin{equation}
    \Psi(\mu) \triangleq \frac{1}{n_H} \sum_{i=1}^{n_H}\psi(\mu, i)
\end{equation}
}
$\Psi$ is implicitly dependent on the dataset so we will compute the average value across all datasets:
\begin{equation}
    \mathrm{Metric1}(\mu) = \sum_{\mathscr{D}} \Psi(\mu; \mathscr{D})
\end{equation}
\textbf{Note}: This metric is only provided for efficient computation since the true metric can be expensive to compute. All ranking will be done using Metric 2 below. In most cases we have tested, metric 1 and metric 2 correlate highly with each other, but metric 2 is the more principled measurement.

\subsubsection{Metric 2: Conditional Mutual Information}
To ensure that a measure is causally informative of generalization, we use a special instance of the Inductive Causation (IC) algorithm to measure whether an edge exists between the complexity measure and observed generalization in a causal probabilistic graph. Specifically, we measure how informative is the complexity measure about generalization when one or more hyper-parameters is observed.

Concretely, we denote $\gO$ as the set of hyper-parameters being conditioned on. For instance, if $\gO = \{\varnothing\}$, then $|\gO| = 0$ the conditional mutual information measures how informative the measure is about generalization in general; if $\gO = \{\text{learning rate}, \text{depth}\}$, then $|\gO| = 2$ the conditional mutual information measures how informative the measure is about generalization when we already know the learning rate and the depth of the model. For a particular $\gO$, we can partition all the models into groups based on their values at the members of $\gO$. For $\gO = \{\Theta_i\}_{i=1}^N$, the groups are models are $\prod_{i=1}^N \Theta_i$. As a concrete example, suppose $\gO = \{\text{learning rate}, \text{depth}\}$ and there are 2 possible learning rates, $\{0.1, 0.01\}$, and 2 depths, $\{8, 16\}$, then there will be 4 groups and models within each group will have the same learning rate and depth.

We further treat $V_\mu$ and $V_g$ as Bernoulli random variables by counting over groups of models. On a particular group $\gO_k$, we can compute:
\begin{equation}
    p(V_g | \gO_k), \quad p(V_\mu | \gO_k), \quad p(V_g, V_\mu | \gO_k)
\end{equation}
These probabilities be easily obtained by counting over the models within $\gO_k$, which further allows us to compute the mutual information between $V_\mu$ and $V_g$ conditioned on $\gO_k$:
\begin{equation}
    \gI(V_g, V_\mu \, | \, \gO_k) = \sum_{V_g} \sum_{V_\mu} p(V_g, V_\mu \, | \, \gO_k) \log\Big(\frac{p(V_\mu, V_g\, | \, \gO_k)}{p(V_\mu\, | \, \gO_k) \, p(V_g\, | \, \gO_k)}\Big)
\end{equation}
Since each $\gO_k$ occurs with equal probability of $p_c = 1\,/\,\prod_{i=1}^N |\Theta_i|$, with slight abuse of notation, we can compute the mutual information between $V_\mu$ and $V_g$ conditioned on that values of $\gO$ is observed as follows:
\begin{equation}
    \gI(V_g, V_\mu \, | \, \gO) = \sum_{\gO_k} p_c \, \gI(V_g, V_\mu \, | \, \gO_k)
\end{equation}
Since here the conditional mutual information between a complexity measure and generalization is at most equal to the conditional entropy of generalization, we normalize it with the conditional entropy to arrive at a criterion ranging between 0 and 1. The conditional entropy of generalization is computed as follows:
\begin{equation}
    \gH(V_g \, | \, \gO) = \sum_{\gO_k} p_c \, \sum_{V_g}  p(V_g\, | \, \gO_k) \log\big(p(V_g\, | \, \gO_k)\big)
\end{equation}
\begin{equation}
   \hat{\gI}(V_g, V_\mu \, | \,\gO) = \frac{\gI(V_g, V_\mu \, | \, \gO)}{\gH(V_g \, | \, \gO)}
\end{equation}
Finally, by the IC algorithm\footnote{For computational practicality, we will often restrict the maximum number of hyper-parameters in $\gO$.}, we take the minimum over all possible $\gO$ and our final metric is the follows:
\begin{equation}
    \gJ(\mu) \triangleq \min_{\gO} \hat{\gI}(V_g, V_\mu \, | \,\gO)
\end{equation}
Similar to $\Psi$, $\gJ(\mu)$ is implicitly dependent on the dataset so we will compute the average value across all datasets:
\begin{equation}
    \mathrm{Metric2}(\mu) = \sum_{\mathscr{D}} \gJ(\mu;\mathscr{D})
\end{equation}
This metric is more principled than metric 1 and it is the \textbf{only} metric for ranking the submissions.

\subsubsection{Human Evaluation}
\prompt{You can include subjective measures provided by human judges (particularly for live /  demonstration competitions). In that case, describe the judging criteria, which must be as orthogonal as possible, sensible, and specific. Provide details on the judging protocol, especially how to break ties between judges. Explain how judges will be recruited and, if possible, give a tentative list of judges, justifying their qualifications.}

Since the competition format is very new, we reserve the rights to inspect the submitted code for abuse (e.g., using the provided compute for tasks unrelated to the competition or tempting with the competition server). We expect the need for human evaluation to be rare.

\subsection{Baselines, code, and material provided}
\prompt{
Specify what are (will be) the baselines for the competition. Provide preliminary results, if available.}

We will be providing baselines in the form of 2 different measures. First measure is the VC-dimension of the models and the second measure is the true generalization gap of the models with added noises. The VC-dimension of convolutional neural networks can be found in \cite{jiang2019fantastic}. The former is meant to be a weak baseline from classical machine learning literature, and the latter is meant to be a strong baseline, which we expect few solution to beat since it is essentially a noisy version of the true quantity of interest.

\prompt{Indicate
whether there will be available code for the participants to get started with (``starting kit''). For certain competitions, material provided may include a hardware platform.}

We will be providing code providing an example measure as well as demonstrating how to access various attribute of the models and how to compute potential values of interest such as norms of the weights or the gradients. Depending on the the feedbacks from the community, we may also add new baselines.

\subsection{Tutorial and documentation}
\prompt{
Provide a reference to a white paper you wrote describing the
problem and/or explain what tutorial material you will provide.}

The inspiration and much backbone of the this competition can be found in \cite{jiang2019fantastic}, which identifies various challenges of evaluating generalization and outlines the rigorous procedure to demonstrate the effectiveness of certain generalization measure. The procedure is modified and reproduce in the metric shown above. We will also prepare a tutorial on generalization in deep learning for participants who are not as familiar with the field, and provide a list of reference for further readings. We will also be providing the API of the software used in the competition.

\section{Organizational aspects}
\subsection{Protocol}

\prompt{
Explain the procedure of the competition: what the participants will have to do, what will be submitted (results or code), and the evaluation procedure.
Will there be several phases? Will you use a competition platform with on-line submissions and a leader board? Indicate means of preventing cheating.
Provide your plan to organize beta tests of your protocol and/or platform.}

The competition will be separated into two phases. In \textbf{development phase} (phase 1) of the competition, the competitors will develop solutions on the public data that we provide, and submit solutions which we will evaluate on the Phase 1 private dataset.
% Between \textbf{Phase 1} and \textbf{Phase 2}, the competitors will have a week to prepare and submit their final solution.
After \textbf{evaluation phase} (phase 2) starts, the competitors are expected to submit their final solution. The final solution will be evaluated on \textbf{Phase 1} data first to check if it finishes within time. If the solution finishes on Phase 1 within time then their code will be run on the Phase 2 private dataset without any time limit. Both private datasets would be models trained on different hyper-parameters from the public dataset. The competitors are expected to submit code which we will evaluate on the cloud. We will be using \textbf{\hyperlink{https://codalab.org/}{Codalab}} to orchestrate the online submissions and provide a live leader board.
At NeurIPS, we will be organizing a workshop for top-performing teams to present their solutions. There will be posters and also oral presentation. We also plan to invite guest speakers. The details of the workshop is still being finalized.

\subsection{Rules}
\prompt{
In this section, please provide:

\begin{itemize}
	\item \textbf{A verbatim copy of (a draft of) the contest rules given to the contestants.}
	\item \textbf{A discussion of those rules and how they lead to the desired outcome of your competition.} 
	\item \textbf{A discussion about cheating prevention.}
\end{itemize}
\textbf{
Choose inclusive rules, which allow the broadest possible participation from the NeurIPS audience.}
}
\subsubsection{Draft of Rules}
\begin{enumerate}
    \item Participants are expected to form teams. There are no limits on number of participants on each team.
    \item Each participant can only be on one team. Violation may result in disqualification.
    \item Since we may use publicly available dataset, including the datasets in the submission for any form of direct lookup is not permitted. Violation may result in disqualification.
    \item Each team need to submit a academic-paper-style write up that describes their solution to be eligible for winning in the evaluation phase.
    \item Top eligible teams will be invited to give a presentation at NeurIPS 2020.
    % \item For theory track, to be considered for winning, the competitors must also submit a write-up which justifies the complexity measure if not provides a generalization bound base on the complexity measure. In this track, the competitors are also not allowed to explicitly use models beyond those provided.
    % \item For free-form track, anything is fair-game. This includes regression analysis base on a large number of models beyond those provided; however, to be considered for winning, a write-up is also required.
    \item All submission in \textbf{Phase 1} are required to finish executing in a fixed time limit; submissions that exceed the computational limit will receive minimum score (0.0). We will announce the time limit soon, but a submission should on average be able to process a model within 5 minutes wall-clock time on GPU. Hardwares specs are currently in preparation.
    \item Only one submission is allowed in \textbf{Phase 2} for final scoring.
        \begin{itemize}
            \item  The submission will be first run on the \textbf{Phase 1} data. If the submission times out or fails on the Phase 1 data, then it will not be evaluated on Phase 2 data; however, the competitors will be allowed to resubmit changed version of the code.
            \item If the submission finishes running within the time limit, the submission will be run on \textbf{Phase 2} datasets without time limit. This process can only be done \textbf{once}, after which any further submission from the team is not evaluated.
        \end{itemize}
    \item Computation resource is allocated on a first-come-first-serve basis. A maximum submissions of 5 is allowed per day for each team. This number is subject to change.
    \item Competitors affiliated with Alphabet and co-organizers are not eligible for winning but participation is allowed.
    % \item \textbf{Incomplete}: Some intellectual property statement. Require the code to be open-sourced or ask them to give us the right to execute the code but the participants keep the ownership? what kind of license are we looking at.
    % \item \textbf{Incomplete}: Who can enter the competition? We want to be inclusive but we are also providing massive amount of compute so we want to do manual approval for entering the contest? What information are we going to ask from the competitors.
    % \item We expect the competitors' code to be open-sourced after the end of the competition; otherwise, the competitor should grant the organizers the rights to run the code.
\end{enumerate}

\subsubsection{Discussion}
Our primary interest is to provide a platform for the community to test out new theories to improve the understanding of why deep neural networks generalize, and rigorously analyze how much these theories reflect the real models. By asking for a write-up on the theoretical motivation behind the proposed complexity measures, we hope to motivate the competitors to build their solutions in a more principled manner. However, solutions that use parametric solutions or black-box methods are also valuable since they tend to be more expressive and powerful. In this case, a write-up will also help practitioners in the community.

We are providing the compute for the evaluation, which both prevent cheating since the competitors do not have access to the data and lower the barrier-to-entry for participants who may not have access to large amounts of compute resources.
The reason for limiting submissions per day is two-fold:
\begin{itemize}
    \item It is unclear if it is possible to overfit to the private dataset since the competition format is extremely new.
    \item We need to restrict the compute for practicality since we are providing free compute and we want to prevent abuse. We will also adjust the time budget based on user feedback.
\end{itemize}

Current proposal of the competition only includes a set of sequential feed-forward models trained to small loss on several image classification benchmarks. In the future iteration of this competition, we plan to support more classes of computational graph (e.g. ResNet) and at different loss values. We are also considering including tracks for transfer learning.

\subsubsection{Cheating Prevention}
The private models and data that will be used in the competitions will be created from scratch so the competitors will not be able to find them online. Further, the submission will not be able to access the internet while the code is running, which prevents the information of the dataset from leaking. We also put compute time limits on the submissions. Solutions that exceed the time limit will receive the minimum score of 0.0. We are only allowing a small number of submissions everyday to minimize the possibility of reverse engineering. Finally, as the last resort, if we observe unusual behavior, we will manually inspect the competitors' source code.

% \subsection{Schedule and readiness}
\prompt{
Provide a time line for competition preparation and for running the competition itself. Propose a reasonable schedule leaving enough time for the organizers
to prepare the event (a few months), enough time for the participants to develop their methods (e.g. 90 days), enough time for the organizers to review the entries, analyze and publish the results. 

For live/demonstration competitions, indicate how much overall time you will need (we do not guarantee all competitions will get the time they request). Also provide a detailed schedule for the on-site contest at NeurIPS. This schedule should at least include times for introduction talks/video presentations, demos by the contestants, and an award ceremony. 

 Will the participants need to prepare their contribution in advance (e.g. prepare a demonstration) and bring ready-made software and hardware to the competition site? Or, on the contrary, can will they be provided with everything they need to enter the competition on the day of the competition? Do they need to register in advance? What can they expect to be available to them on the premises of the live competition (tables, outlets, hardware, software and network connectivity)? What do they need to bring (multiple connectors, extension cords, etc.)?

Indicate what, at the time of writing this proposal, is already ready.}

% Apr 13: Know whether approved.
% + 2 months
% Jun 13: Start beta.
% Aug 01: Start real competition
% Oct 01: Close submissions, 
\section{Timeline}
\begin{itemize}
    \item \textbf{Jul 15}: \textbf{Phase 1 starts}. Participants submit their solution to be evaluated on the Phase 1 private dataset.
    % \item \textbf{Oct 01}: \textbf{Phase 1 ends}. Participants submit their solutions for phase 2.
    \item \textbf{Oct 08}: \textbf{Phase 2 begins}. Participants' code are evaluated on the Phase 2 private dataset.
    \item \textbf{Oct 24}: \textbf{Phase 2 ends}. All computation finalized.
    \item \textbf{Oct 31}: Results are announced.
    \item \textbf{Dec 11}: Winning teams are invited to present at the conference
\end{itemize}

\section{Organizing team}
\begin{itemize}
    \item \textbf{Yiding Jiang}: Yiding Jiang is an AI resident at Google Research. He previously received Bachelor of Science in Electrical Engineering and Computer Science from University of California, Berkeley. He has worked on projects in deep learning, reinforcement learning and robotics. He has published papers related to predicting generalization of neural networks~\cite{jiang2018predicting, song2019observational} and evaluating complexity measures~\cite{jiang2019fantastic}.
    \item \textbf{Pierre Foret}: Pierre Foret is an AI resident at Google Research. He previously received a Master of Financial Engineering from University of California, Berkeley and a Master in applied math from ENSAE Paristech. His research interests lie in the intersection of optimization and generalization in deep learning.
    \item \textbf{Scott Yak}: Scott Yak is a Software Engineer at Google Research. He has previously received a Bachelor of Science and Engineering at Princeton University. He is currently working on AutoML and Neural Architecture Search at Google. He has published work on predicting generalization of neural networks~\cite{yak2019towards}.
    \item \textbf{Behnam Neyshabur}: Behnam Neyshabur is a senior research scientist at Google. Before that, he was a postdoctoral researcher at New York University and a member of Theoretical Machine Learning program at Institute for Advanced Study (IAS) in Princeton. In summer 2017, He received a PhD in computer science at TTI-Chicago. He is interested in machine learning and optimization and his primary research is on optimization and generalization in deep learning. He has co-organized ICML 2019 workshops on ``Understanding and Improving Generalization in Deep Learning'' and ``Identifying and Understanding Deep Learning Phenomen''. He has published several papers related to complexity measures and generalization in deep learning~\cite{jiang2019fantastic,song2019observational,chatterji2019intriguing,neyshabur2015search,neyshabur2015norm,neyshabur2017geometry,neyshabur2017exploring,neyshabur2018pac,neyshabur2017implicit,arora18b,neyshabur2018towards}
    \item \textbf{Hossein Mobahi}: Hossein Mobahi is a research scientist at Google Research. His recent efforts covers the intersection of machine learning, generalization and optimization, with emphasis on deep learning. Prior to joining Google in 2016, he was a postdoctoral researcher in the Computer Science and Artificial Intelligence Lab (CSAIL) at MIT. He obtained his PhD in Computer Science from the University of Illinois at Urbana-Champaign (UIUC). He is the recipient of Computational Science \& Engineering Fellowship, Cognitive Science \& AI Award, and Mavis Memorial Scholarship. He has published several works on generalization~\cite{jiang2019fantastic, jiang2018predicting, elsayed2018large} and theoretical foundations of self-distillation~\cite{mobahi2020self}.
    \item \textbf{Gintare Karolina Dziugaite}: Gintare Karolina Dziugaite is a Fundamental Research Scientist at Element AI. Dziugaite recently graduated from the University of Cambridge, where she completed her doctorate in Zoubin Ghahramani’s machine learning group. The focus of her thesis was on constructing generalization bounds to understand existing learning algorithms in deep learning and propose new ones. She continues to work on explaining generalization phenomenon in deep learning using statistical learning tools. She was a lead organizer for the 2019 ICML workshop on Machine Learning with Guarantees. She has published several works in generalization for deep learning~\cite{dziugaite2017computing, dziugaite2018data, negrea2019information, dziugaite2020revisiting}.
    % Karolina is a Fundamental Research Scientist at Element AI. Her research combines theoretical and empirical approaches to understanding deep learning, with a focus on generalization and network compression. Before joining Element, She obtained her Ph.D. in machine learning from the University of Cambridge, under the supervision of Zoubin Ghahramani. She studied Mathematics at the University of Warwick and read Part III in Mathematics at the University of Cambridge, receiving a Masters of Advanced Study (MASt) in Applied Mathematics. She has published several works in generalization for deep learning~\cite{dziugaite2017computing, dziugaite2018data, negrea2019information, dziugaite2020revisiting}.
    \item \textbf{Daniel M. Roy}: Daniel M. Roy is an Assistant Professor in the Department of Statistical Sciences at the University of Toronto and Canada CIFAR AI Chair.  Prior to joining Toronto, Roy was a Research Fellow of Emmanuel College and Newton International Fellow of the Royal Society and Royal Academy of Engineering, hosted by the University of Cambridge. Roy completed his doctorate in Computer Science at the Massachusetts Institute of Technology. Roy has co-organized a number of workshops, including the 2008, 2012, and 2014 NeurIPS Workshops on Probabilistic Programming, a 2016 Simons Institute Workshop on Uncertainty in Computation, and special sessions in 2019 at the Statistical Society of Canada meeting and 2016 at the Mathematical Foundations of Programming Semantics conference. Last year at ICML, he organized a workshop on Machine Learning with Guarantees. He has published several works in generalization for deep learning~\cite{yang2019fast, dziugaite2017computing, dziugaite2018data, negrea2019information}.
    % Daniel is an assistant professor at the Department of Statistical Sciences and Department of Computer Science of University of Toronto. Daniel’s research blends computer science, statistics and probability theory; He studies “probabilistic programming” and develop computational perspectives on fundamental ideas in probability theory and statistics. Daniel is particularly interested in: representation theorems that connect computability, complexity, and probabilistic structures; stochastic processes, the use of recursion to define stochastic processes, and applications to nonparametric Bayesian statistics; and the complexity of probabilistic and statistical inference, especially in the context of probabilistic programming. Ultimately, Daniel is motivated by the long term goal of making lasting contributions. He has published several foundational works in generalization for deep learning~\cite{yang2019fast, dziugaite2017computing, dziugaite2018data, negrea2019information}.
    \item \textbf{Suriya Gunasekar}: Suriya Gunasekar is a senior researcher at the Machine Learning and Optimization (MLO) Group of Microsoft Research. Prior to joining MSR, she was a Research Assistant Professor at Toyota Technological Institute at Chicago. She received my PhD in ECE from The University of Texas at Austin. She has published several works in optimization and implicit regularization~\cite{soudry2018implicit, gunasekar2017implicit, gunasekar2018implicit, gunasekar2018characterizing}.
    \item \textbf{Isabelle Guyon}: Isabelle Guyon is chaired professor in “big data” at the Université Paris-Saclay, specialized in statistical data analysis, pattern recognition and machine learning. She is one of the cofounders of the ChaLearn Looking at People (LAP) challenge series and she pioneered applications of the MIcrosoft Kinect to gesture recognition. Her areas of expertise include computer vision and and bioinformatics. Prior to joining ParisSaclay she worked as an independent consultant and was a researcher at AT\&T Bell Laboratories, where she pioneered applications of neural networks to pen computer interfaces (with collaborators including Yann LeCun and Yoshua Bengio) and coinvented with Bernhard Boser and Vladimir Vapnik Support Vector Machines (SVM), which became a textbook machine learning method. She worked on early applications of Convolutional Neural Networks (CNN) to handwriting recognition in the 1990’s. She is also the primary inventor of SVMRFE, a variable selection technique based on SVM. The SVMRFE paper has thousands of citations and is often used as a reference method against which new feature selection methods are benchmarked. She also authored a seminal paper on feature selection that received thousands of citations. She organized many challenges in Machine Learning since 2003 supported by the EU network Pascal2, NSF, and DARPA, with prizes sponsored by Microsoft, Google, Facebook, Amazon, Disney Research, and Texas Instrument. Isabelle Guyon holds a Ph.D. degree in Physical Sciences of the University Pierre and Marie Curie, Paris, France. She is president of Chalearn, a nonprofit dedicated to organizing challenges, vice president of the Unipen foundation, adjunct professor at NewYork University, action editor of the Journal of Machine Learning Research, editor of the Challenges in Machine Learning book series of Microtome, and program chair of the upcoming NIPS 2016 conference.
    \item \textbf{Samy Bengio}: Samy Bengio (PhD in computer science, University of Montreal, 1993) is a research scientist at Google since 2007. He currently leads a group of research scientists in the Google Brain team, conducting research in many areas of machine learning such as deep architectures, representation learning, sequence processing, speech recognition, image understanding, large-scale problems, adversarial settings, etc. He was the general chair for Neural Information Processing Systems (NeurIPS) 2018, the main conference venue for machine learning, was the program chair for NeurIPS in 2017, is action editor of the Journal of Machine Learning Research and on the editorial board of the Machine Learning Journal, was program chair of the International Conference on Learning Representations (ICLR 2015, 2016), general chair of BayLearn (2012-2015) and the Workshops on Machine Learning for Multimodal Interactions (MLMI'2004-2006), as well as the IEEE Workshop on Neural Networks for Signal Processing (NNSP'2002), and on the program committee of several international conferences such as NIPS, ICML, ICLR, ECML and IJCAI. 
\end{itemize}

\prompt{
Provide a short biography of all team members, stressing their competence for their assignments in the competition organization. Please note that diversity in the organizing team is encouraged, please elaborate on this aspect as well.  Make sure to include: coordinators, data providers, platform administrators, baseline method providers, beta testers, and evaluators. }
\section{Resources}
\begin{itemize}
    \item \textbf{Community} We will facilitate a public forum where the organizers interact with the participants and answer any potential questions. We hope the platform would reduce communication friction and also foster a community among the competitors.
    \item \textbf{Computing Resources} We will provide all the computational resources needed for evaluating the complexity measures. We will also be providing learning resources for those not as knowledgeable about statistical learning theory and generalization in deep learning.
    \item \textbf{Prize} None.
\end{itemize}

\prompt{Describe your resources (computers, support staff, equipment, sponsors, and available prizes and travel awards).

For live/demonstration competitions, explain how much will be provided by the organizers (demo framework, software, hardware) and what the participants will need to contribute (laptop, phone, other hardware or software).}

% \subsection{Support and facilities requested}
% \prompt{
% Please indicate the kind of support and facilities you need from the conference. 

% For live/demonstration competitions, indicate what you will need on site from NeurIPS in order to run the live competition (size of room, duration, tables, outlets, network connectivity). We do not commit to provide all such support free-of-charge. }

% We would like to request a small to medium size room (100 to 150 audiences) collocated with the workshops for the winners of the competition to present their solutions. The presentations and invited talks would last for 3 to 4 hours, and we will need network connectivity and projector for the presentation and potential remote presentations. We would also like to request support for poster presentations (i.e. poster stands or large area of walls).

% We are currently exploring the possible options for travel grant support. However, since we are providing constant computing resource through out the duration of entire competition, and, given the novelty of the competition format, we do not have a good estimate of the number of teams, we would like to ask the organizers of NeurIPS to give winners of the competition that are from underrepresented groups higher priority for various applicable travel grants.

\bibliographystyle{acm} \bibliography{biblio}

\begin{thebibliography}{10}

\bibitem{arora18b}
{\sc Arora, S., Ge, R., Neyshabur, B., and Zhang (Alphabetical~Order), Y.}
\newblock Stronger generalization bounds for deep nets via a compression
  approach.
\newblock In {\em Proceedings of the 35th International Conference on Machine
  Learning (ICML)\/} (2018), pp.~254--263.

\bibitem{chatterji2019intriguing}
{\sc Chatterji, N.~S., Neyshabur, B., and Sedghi, H.}
\newblock The intriguing role of module criticality in the generalization of
  deep networks.
\newblock In {\em International Conference on Learning Representations
  (ICLR)\/} (2020 {\bf(spotlight)}).

\bibitem{cinic10}
{\sc Darlow, L.~N., Crowley, E.~J., Antoniou, A., and Storkey, A.~J.}
\newblock Cinic-10 is not imagenet or cifar-10, 2018.

\bibitem{dziugaite2020revisiting}
{\sc Dziugaite, G.~K.}
\newblock {\em Revisiting Generalization for Deep Learning: PAC-Bayes, Flat
  Minima, and Generative Models}.
\newblock PhD thesis, University of Cambridge, 2020.

\bibitem{dziugaite2017computing}
{\sc Dziugaite, G.~K., and Roy, D.~M.}
\newblock Computing nonvacuous generalization bounds for deep (stochastic)
  neural networks with many more parameters than training data.
\newblock {\em arXiv preprint arXiv:1703.11008\/} (2017).

\bibitem{dziugaite2018data}
{\sc Dziugaite, G.~K., and Roy, D.~M.}
\newblock Data-dependent pac-bayes priors via differential privacy.
\newblock In {\em Advances in Neural Information Processing Systems\/} (2018),
  pp.~8430--8441.

\bibitem{elsayed2018large}
{\sc Elsayed, G., Krishnan, D., Mobahi, H., Regan, K., and Bengio, S.}
\newblock Large margin deep networks for classification.
\newblock In {\em Advances in neural information processing systems\/} (2018),
  pp.~842--852.

\bibitem{2018arXiv180303635F}
{\sc {Frankle}, J., and {Carbin}, M.}
\newblock {The Lottery Ticket Hypothesis: Finding Sparse, Trainable Neural
  Networks}.
\newblock {\em arXiv e-prints\/} (Mar 2018), arXiv:1803.03635.

\bibitem{gunasekar2018characterizing}
{\sc Gunasekar, S., Lee, J., Soudry, D., and Srebro, N.}
\newblock Characterizing implicit bias in terms of optimization geometry.
\newblock {\em arXiv preprint arXiv:1802.08246\/} (2018).

\bibitem{gunasekar2018implicit}
{\sc Gunasekar, S., Lee, J.~D., Soudry, D., and Srebro, N.}
\newblock Implicit bias of gradient descent on linear convolutional networks.
\newblock In {\em Advances in Neural Information Processing Systems\/} (2018),
  pp.~9461--9471.

\bibitem{gunasekar2017implicit}
{\sc Gunasekar, S., Woodworth, B.~E., Bhojanapalli, S., Neyshabur, B., and
  Srebro, N.}
\newblock Implicit regularization in matrix factorization.
\newblock In {\em Advances in Neural Information Processing Systems\/} (2017),
  pp.~6151--6159.

\bibitem{2015arXiv150602626H}
{\sc {Han}, S., {Pool}, J., {Tran}, J., and {Dally}, W.~J.}
\newblock {Learning both Weights and Connections for Efficient Neural
  Networks}.
\newblock {\em arXiv e-prints\/} (Jun 2015), arXiv:1506.02626.

\bibitem{jiang2018predicting}
{\sc Jiang, Y., Krishnan, D., Mobahi, H., and Bengio, S.}
\newblock Predicting the generalization gap in deep networks with margin
  distributions.
\newblock {\em arXiv preprint arXiv:1810.00113\/} (2018).

\bibitem{jiang2019fantastic}
{\sc Jiang, Y., Neyshabur, B., Krishnan, D., Mobahi, H., and Bengio, S.}
\newblock Fantastic generalization measures and where to find them.
\newblock In {\em International Conference on Learning Representations\/}
  (2019).

\bibitem{cifar10}
{\sc Krizhevsky, A.}
\newblock Learning multiple layers of features from tiny images.
\newblock Tech. rep., 2009.

\bibitem{lin2013network}
{\sc Lin, M., Chen, Q., and Yan, S.}
\newblock Network in network.
\newblock {\em arXiv preprint arXiv:1312.4400\/} (2013).

\bibitem{2018arXiv181005270L}
{\sc {Liu}, Z., {Sun}, M., {Zhou}, T., {Huang}, G., and {Darrell}, T.}
\newblock {Rethinking the Value of Network Pruning}.
\newblock {\em arXiv e-prints\/} (Oct 2018), arXiv:1810.05270.

\bibitem{mobahi2020self}
{\sc Mobahi, H., Farajtabar, M., and Bartlett, P.~L.}
\newblock Self-distillation amplifies regularization in hilbert space.
\newblock {\em arXiv preprint arXiv:2002.05715\/} (2020).

\bibitem{negrea2019information}
{\sc Negrea, J., Haghifam, M., Dziugaite, G.~K., Khisti, A., and Roy, D.~M.}
\newblock Information-theoretic generalization bounds for sgld via
  data-dependent estimates.
\newblock In {\em Advances in Neural Information Processing Systems\/} (2019),
  pp.~11013--11023.

\bibitem{svhn}
{\sc Netzer, Y., Wang, T., Coates, A., Bissacco, A., Wu, B., and Ng, A.~Y.}
\newblock Reading digits in natural images with unsupervised feature learning.

\bibitem{neyshabur2017implicit}
{\sc Neyshabur, B.}
\newblock {\em Implicit Regularization in Deep Learning}.
\newblock PhD thesis, TTIC, 2017.

\bibitem{neyshabur2017exploring}
{\sc Neyshabur, B., Bhojanapalli, S., McAllester, D., and Srebro, N.}
\newblock Exploring generalization in deep learning.
\newblock In {\em Advances in Neural Information Processing Systems (NIPS)\/}
  (2017), pp.~5947--5956.

\bibitem{neyshabur2018pac}
{\sc Neyshabur, B., Bhojanapalli, S., and Srebro, N.}
\newblock A pac-bayesian approach to spectrally-normalized margin bounds for
  neural networks.
\newblock In {\em International Conference on Learning Representations
  (ICLR)\/} (2018).

\bibitem{neyshabur2018towards}
{\sc Neyshabur, B., Li, Z., Bhojanapalli, S., LeCun, Y., and Srebro, N.}
\newblock Towards understanding the role of over-parametrization in
  generalization of neural networks.
\newblock In {\em International Conference on Learning Representations
  (ICLR)\/} (2019).

\bibitem{neyshabur2017geometry}
{\sc Neyshabur, B., Tomioka, R., Salakhutdinov, R., and Srebro, N.}
\newblock Geometry of optimization and implicit regularization in deep
  learning.
\newblock {\em arXiv preprint arXiv:1705.03071\/} (2017).

\bibitem{neyshabur2015search}
{\sc Neyshabur, B., Tomioka, R., and Srebro, N.}
\newblock In search of the real inductive bias: On the role of implicit
  regularization in deep learning.
\newblock In {\em International Conference on Learning Representations (ICLR)
  workshop\/} (2015).

\bibitem{neyshabur2015norm}
{\sc Neyshabur, B., Tomioka, R., and Srebro, N.}
\newblock Norm-based capacity control in neural networks.
\newblock In {\em Conference on Learning Theory (COLT)\/} (2015),
  pp.~1376--1401.

\bibitem{oxfordflowers}
{\sc Nilsback, M.-E., and Zisserman, A.}
\newblock Automated flower classification over a large number of classes.
\newblock In {\em Proceedings of the Indian Conference on Computer Vision,
  Graphics and Image Processing\/} (Dec 2008).

\bibitem{oxfordpets}
{\sc Parkhi, O.~M., Vedaldi, A., Zisserman, A., and Jawahar, C.~V.}
\newblock Cats and dogs.
\newblock In {\em IEEE Conference on Computer Vision and Pattern Recognition\/}
  (2012).

\bibitem{Simonyan15}
{\sc Simonyan, K., and Zisserman, A.}
\newblock Very deep convolutional networks for large-scale image recognition.
\newblock In {\em International Conference on Learning Representations\/}
  (2015).

\bibitem{song2019observational}
{\sc Song, X., Jiang, Y., Du, Y., and Neyshabur, B.}
\newblock Observational overfitting in reinforcement learning.
\newblock In {\em International Conference on Learning Representations
  (ICLR)\/} (2020).

\bibitem{soudry2018implicit}
{\sc Soudry, D., Hoffer, E., Nacson, M.~S., Gunasekar, S., and Srebro, N.}
\newblock The implicit bias of gradient descent on separable data.
\newblock {\em The Journal of Machine Learning Research 19}, 1 (2018),
  2822--2878.

\bibitem{fashionmnist}
{\sc Xiao, H., Rasul, K., and Vollgraf, R.}
\newblock Fashion-mnist: a novel image dataset for benchmarking machine
  learning algorithms.
\newblock {\em CoRR abs/1708.07747\/} (2017).

\bibitem{yak2019towards}
{\sc Yak, S., Gonzalvo, J., and Mazzawi, H.}
\newblock Towards task and architecture-independent generalization gap
  predictors.
\newblock {\em arXiv preprint arXiv:1906.01550\/} (2019).

\bibitem{yang2019fast}
{\sc Yang, J., Sun, S., and Roy, D.~M.}
\newblock Fast-rate pac-bayes generalization bounds via shifted rademacher
  processes.
\newblock In {\em Advances in Neural Information Processing Systems\/} (2019),
  pp.~10802--10812.

\end{thebibliography}

\end{document}